[Title Page]

# HFT-ONLSTM: Hierarchical and Fine-Tuning Multi-label Text Classification

Pengfei Gao[1], Jingpeng Zhao[1], Yinglong Ma[1,*], Tanvir Ahmad[1], Beihong Jin[2]

1. School of Control and Computer Engineering, North China Electric Power University, Beijing 102206, China.

   Email: Gau@ncepu.edu.cn, 1182227090@ncepu.edu.cn, yinglongma@ncepu.edu.cn，Tanvirahmad@ncepu.edu.cn

2. Institute of Software, Chinese Academy of Sciences, Beijing 100190, China

   Email: beihongjin@iscas.cn

\*  Correspondence author

Correspondence information:

Full name: Yinglong Ma

Affiliation:   School of Control and Computer Engineering, North China Electric Power University, Beijing 102206, China,

Email address: yinglongma@ncepu.edu.cn

Telephone number: +86 10 61772643

HFT-ONLSTM: Hierarchical and Fine-Tuning Multi-label Text Classification

**Abstract**

Many important classification problems in the real-world consist of a large number of closely related categories in a hierarchical structure or taxonomy. Hierarchical multi-label text classification (HMTC) with higher accuracy over large sets of closely related categories organized in a hierarchy or taxonomy has become a challenging problem. In this paper, we present a hierarchical and fine-tuning approach based on the Ordered Neural LSTM neural network, abbreviated as HFT-ONLSTM, for more accurate level-by-level HMTC. First, we present a novel approach to learning the joint embeddings based on parent category labels and textual data for accurately capturing the joint features of both category labels and texts. Second, a fine tuning technique is adopted for training parameters such that the text classification results in the upper level should contribute to the classification in the lower one. At last, the comprehensive analysis is made based on extensive experiments in comparison with the state-of-the-art hierarchical and flat multi-label text classification approaches over two benchmark datasets, and the experimental results show that our HFT-ONLSTM approach outperforms these approaches, in particular reducing computational costs while achieving superior performance.

## 1    Introduction

As the number of textual documents drastically increases, many important classification problems in the real-world consist of a large number of categories that are ordinarily quite close and can be organized in a hierarchical structure or taxonomy. Web directories (e.g, The Open Directory Project/DMOZ[1]), medical classification schemes (e.g., Medical Subject Headings[2]), the library and patent classification scheme[3], the Wikipedia topic classifications[4], and social media websites [45], etc., are some of the typical examples of large hierarchical text repositories. The urge of automatic text classification as well as more refinement methods needs to catch up with these needs to provide more accurate and reliable solution.

---

[1] https://www.dmoz-odp.org/
[2] https://meshb.nlm.nih.gov/treeView
[3] https://www.loc.gov/aba/cataloging/classification/
[4] https://en.wikipedia.org/wiki/Portal:Contents/Categories

Text classification (TC) is intended to classify textual data automatically into categories and has been commonly used for automatic email categorization as well as detection of spam, etc. In general, a typical TC performs well in a situation when there are only two or a few numbers of well-separated categories, but it's become ineffective while dealing with a large number of categories/classes [1]. The problem of text classification with greater precision over broad sets of closely related categories is fundamentally difficult [2]. This is especially true for accurate text classification in some large hierarchical text repositories mentioned above.

The hierarchical classification (HC) is a sort of more complex classification function in which categories (classes) are not disassembled but organized into a hierarchical structure [3]. In multi-label classification [2, 13, 36, 37], an entity can belong to one, more than one, or no classes at all. Hierarchical multi-label text classification (HMTC) [2, 3] have issues with a large number of typically very similar categories in a hierarchical system where a piece of text will correspond to one or more taxonomic hierarchical nodes. Each category corresponds to a node in the taxonomic hierarchy, and all categories are closely related in terms of the hierarchical structure. Many flat classifiers actually disregard the hierarchical structure by "flattening" it to the level of the leaf nodes for the classification of multi-label text, and therefore are subject to the problems similar to traditional TC. Undoubtedly, the hierarchical structure information is crucial for building efficient HMTC algorithms to improve text classification accuracy in a case when it comes to handling a large number of categories/classes and attributes.

Recently, HMTC typically has two main kinds of approaches [2, 3], which can be described as global approaches and local approaches. In local approaches, a unique classifier is created for every single node or every level in the taxonomy, whereas global strategies establish a common classification for the entire taxonomy. First, the state-of-the-art of local approaches for HMTC is HDLTex [4], which demonstrate higher accuracy performance over non-neural models. However, local approaches often have relatively higher computation costs and suffer from the problem of tremendous parameter variables possibly causing parameter explosion (similar to other local approaches). The state-of-the-art global approach for HMTC is based on a unified global deep neural network model [5], where a hierarchical neural attention based text classifier (HATC) relieves the model parameter explosion problem. Unfortunately, HATC typically suffers from the inherent disadvantage of global approaches: the built classifier is not robust enough to account for changes in the composition of the categories [6]. Second, many experimental observations including HATC and HDLTex have illustrated that the accuracy of many approaches for HMTC are not always better than that of some flat classifiers [2, 7]. We argue that the inner information residing in category hierarchy or taxonomies, e.g., semantic association between different levels in the hierarchy, should be carefully analyzed and HMTC should be fine-tuned with the aid of these inner information for classification with high accuracy. At last, most of existing approaches for HMTC often use some neural network models such as CNN and LSTM, etc. These models are often suitable for learning a chain structure on the text, but they are difficult to deal with the hierarchically structured text analysis, especially for the situation where smaller units (e.g., sentences) reside in larger units (e.g., paragraphs). So some new models should be adopted to actually cater for the hierarchically structured text for HMTC.

In this paper, we present a hierarchical and fine-tuning deep learning approach based on the Ordered Neural LSTM model, abbreviated as HFT-ONLSTM, for more accurate HMTC. The contributions of this paper are as follows.

First, we propose a hierarchical and fine-tuning approach (HFT-ONLSTM) for more accurate level-by-level HMTC.

Second, we present a novel approach to learn the joint embeddings based on category labels and textual data. The category labels in the hierarchy and the words in the text are jointly embedded to leverage the hierarchical relations in the hierarchical structure of categories and the textual data for accurately capturing the joint features of both category labels and texts.

Third, a fine tuning technique with respect to the category hierarchy is applied to the Ordered Neural LSTM (ONLSTM) neural network [8] in a level-by-level manner such that the text classification results in the upper level should contribute to the classification in the lower one.

At last, the comprehensive analysis is made based on extensive experiments in comparison with the state-of-the-art hierarchical and flat multi-label text classification approaches over two benchmark datasets, which shows that our HFT-ONLSTM approach outperforms these state-of-the-art approaches, in particular reducing computational costs while achieving superior performance.

This paper is organized as follows. Section 1 is the Introduction. In Section 2, we discuss the related work about HMTC. In section 3, we briefly give the overview of our HMTC approach, and describe how the words and parent categories are jointly embedded and how the fine tuning is made in detail. In Section 4, extensive experiments were made over two textual datasets in comparison with the state-of-the-art multi-label approaches for illustrating the effectiveness and efficiency of our approach. Section 5 is the conclusion.

## 2    Related Work

Currently, the existing deep learning based text classification approaches [9-12] compared with previous machine learning algorithms, have achieved superior efficiency [13, 14] in text classification. HMTC based on deep learning has achieved drastic development and received more attention in recent years, but HMTC's main paradox is how the hierarchical structural relationship of categories can be best used to boost classification efficiency, in contrast to the traditional flat classifiers.

In the last decades, HMTC has mainly covered two aspects [2, 3]: local approaches and global approaches. The typical global HMTC approaches [15-18] use flat model as their backbone, which relies on humans for manual input. These approaches use only training sets on which the entire single classification model is built, which consider all the hierarchical structural categories at once, this makes it complex [3, 6]. While the benefits of using a single global model are that all the parameters for all categories are relatively lower than that of local classifications models. However, since the amount of training data per group at a lower level is considerably smaller than at a higher level, the discriminant features for the parent categories may not be discriminable in the subcategories. In this situation, it is typically difficult for global approaches to use distinct feature sets in multiple category groups. In addition, the global HTMC models may not be sufficiently versatile to account for changes in the hierarchical structure of categories [3, 6].

The typical local HMTC approaches [19-23] mostly use the hierarchical structure for constructing classifiers which are based on local information of the hierarchy. In a top-down manner, the local HMTC approaches can be subdivided into local classifiers per node (LCN), local classifiers per parent node (LCPN) and local classifiers per level (LCL) based on local information used during training phase. For each child node, LCN trains a binary classifier, while LCPN trains a parent node classifier for multi-classes. LCL is used to train a multi-class classifier for the whole hierarchy level. What the top-down local approaches take is basically a tactic to prevent class-level contradictions while using a local hierarchical classifier at the testing phase [3]. The disadvantage of local approaches is that the error propagates from the higher levels to the lower ones. When the classifiers go deeply into the hierarchy, these error propagations make a serious performance loss [24].

The state-of-the-art approaches for HMTC are HDLTex [4], HATC [5] and HFT-CNN [25]. HDLTex and HATC report that they successfully achieve the better performance outperforming most of the existing multi-label classification methods. HDLTex builds a standalone neural network based on either CNN or RNN backbone at each parent node for classification of its children categories, but it needs numerous parameter variables and consumes higher computation cost. HATC is made on an end-to-end global natural attention model, which sequentially predicts the category labels of the next level based on a variant of an attention mechanism [26]. It needs much less parameters than HDLTex for learning an optimal neural network model. HFT-CNN is originally to focus on HMTC for short texts. It uses word embedding and the convolutional layer as parameters to learn the next level of the category hierarchy. What is worth noting is that a fine tuning technique is adopted in which the upper level data contributes to the lower level in categorization. However, the state-of-the-art HMTC approaches including HDLTex, HATC and HFT-CNN are often suitable for learning a chain structure on the text, and therefore are difficult to deal with the hierarchically structured text analysis, especially for the situation where sentences reside in paragraphs. We believe that the performance of the state-of-the-art HMTC approaches is likely to be improved if some hierarchical word representation methods with excellent performance, such as ONLSTM [8], FOREST [38], Sentence-BERT [42] and Tree2Vector [44], etc., can be used to learn word representation in a hierarchy. Furthermore, the state-of-the-art HMTC approaches fail to learn the word semantics of category labels although they do take into account the categorization information at different nodes or levels in the hierarchy during HMTC. We argue that learning the word semantics of category labels together with the hierarchical word representations in the text will be helpful in improving the performance of HMTC.

The main differences between these approaches and our model are as folows. First, we simultaneously learn the word semantics of category labels and the structural semantics of the hierarchy/taxonomy for HMTC, instead of only hierarchical structural semantics. Second, we use the joint embeddings of parent categories and textual data to make the text more discriminable in the subcategories, which can relieve the error propagates from the higher levels to the lower ones. Our work in this paper is based on the extension of our previous work [39, 40]. In [39], we presented a proposal by using a fine-tuning based approach for HMTC and made some preliminary experiments in comparison with two hierarchical text classifiers. In [40], our fine tuning based HMTC approach is further compared with some flat multi-label text classifiers and hierarchical text classifiers. In this paper, we will design and implement our HFT-ONLSTM model that refines the previous fine-tuning based approach in detail. We specifically describe not only the level-by-level text classification based on the fine-tuning technique, but also the training process and parameter transfer within every single level of the hierarchy. Moreover, the deeper analysis of more experimental results is made when the new baseline classifiers are introduced in our experiments for comparison with HFT-ONLSTM. Some new analysis results were obtained based on the comprehensive and extensive experiments, which incur some interesting conclusions mildly different from the previous work.

## 3    The HFT-ONLSTM Approach for HMTC

### 3.1    Hierarchical Categories

In this paper, the proposed HFT-ONLSTM approach deals with the hierarchy or taxonomy where parent-child relationships exist between a category at the upper level and some categories at the next lower level. A parental category possibly has several subcategories, while a children category often has a few ancestral categories including its parental category. Each level in a hierarchical structure generally contains multiple categories with the exception to the root level (just one category).

Let $H = (C, E, r, \tau)$ be a category hierarchy or taxonomy, where $C = \bigcup_{l=1}^{L} C^l$ is the set of category nodes, $C^l$ represents the set of category nodes at the *l-th* level, and *L* is the number of levels in *H*, and $r \in C$ is the root category node. $E \subseteq C^l \times C^{l+1}$ is the set of parent-child relationships for each $1 \leq l \leq L-1$, where *L* be the total number of levels in the hierarchy *H*. Especially, $C^1$ just includes one category node *r*. For any $c \in C$, $\tau(c)$ is the category label of node *c*. A category label can be regarded as a short text with multiple words.

A typical dataset with a hierarchical category structure is DBpedia[5], which includes a large multi-domain category structure derived from Wikipedia as well as other localized versions. Figure 1 is an example of partial segments about the hierarchical categories in the dataset DBpedia that has three hierarchical levels and was first used in the literature [12] for flat text classification.

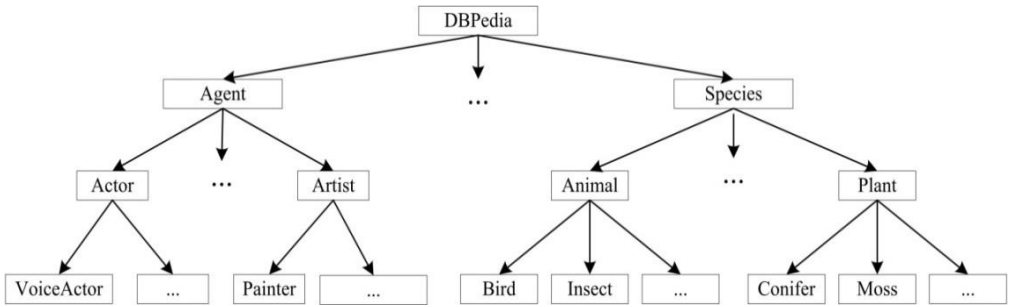

**Fig. 1.** The hierarchical categories in DBpedia.

### 3.2 Level-by-Level Classification

In this paper, we shared the concept about hierarchical multi-label text classification that represents multi-label text classification in a hierarchical manner or learns hierarchical structure for multi-label text classification, just like the existing works such as [4] and [25]. In [4], a binary classifier (a special case of multi-class classification) is applied to each node label with a parent-child hierarchical relation for multi-label classification. In [25], the CNN-based multi-class classifier is used to learn hierarchical structure for multi-label text classification in a level-by-level manner. Our hierarchical multi-label text classification is reduced to finding a set of labels for the text in each path from the root to the leaf [41].

Our HFT-ONLSTM approach adopts a level-by-level hierarchical approach to multi-label text classification. The text is continually classified at each level in the hierarchy. For classification at each level, the text is classified into one category label from the categories at the current level. For the whole process of multi-label classification, the maximum number of category labels that the text is associated with is just the number of levels in the hierarchy.

The framework of the level-by-level multi-label text classification is shown in the

---

[5] https://wiki.dbpedia.org/services-resources/datasets/dbpedia-datasets

HFT-ONLSTM: Hierarchical and Fine-Tuning Multi-label Text Classification

following Figure 2. The notation $\tau(C_i^{j-1})$ refers to the predicted parent category label of the current text $x_i$ at the (*j-1*)-th level, where $j \geq 2$. Here, we use the notation $\alpha \oplus \beta$ to represent that the parent category label $\alpha$ predicted from the upper level is concatenated with the currently corresponding text $\beta$. The notation *w_i* is the word embedding of the *i-th* word of the text, and $h_i$ is the state output based on ONLSTM. The notation *W* denotes the training parameters from the upper level training that will be transferred into the training model at the next level by using fine-tuning technique.

Specifically speaking, for the classification at the *j-th* level, we first combine the corresponding text with its predicted parent category label at the (*j-1*)-*th* level by a text concatenation operation. The concatenated text is further represented in the same vector space by using word embedding matrix. What is worth noting that since that the root category does not have parent category label so there is no need to embed the parent category label. Formally, when level $j = 1$, we consider $\tau(C_i^{j-1})=\tau(C_i^0)= \emptyset$.

After this, what we do is to pass the joint embedded parent category label and text vector representation to deep learning model for training. The deep learning based model used in this paper can be described in two parts: the ONLSTM models and the second one is a multi-layer perceptron (MLP). While training the model, the parameters of training model related to ONLSTM in the upper level are transferred to the lower level, and are further used to fine tune the parameters training of the current level to ensure that the upper level parameters lead to a more precise classification at the lower level adjacent to them. The same process is repeated until the bottom level categories are trained.

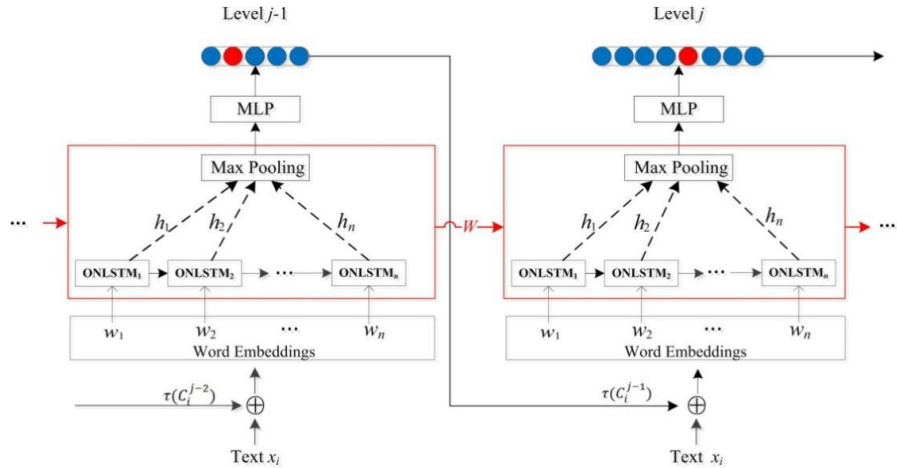

**Fig. 2.** The framework based on level-by-level multi-label classification.

## 3.3 Joint Embedding Based on Parent Category Label and Text

In a hierarchy of taxonomic categories, a parent category can be associated with one or more subcategories, while a subcategory corresponds to only one parent category. There is the conceptual inclusion relationship between a parent category and all its child categories, i.e., the subcategory text must also belong to its parent category. In some hierarchical text classification approaches, such as HATC [5], etc., the classification result of the upper level has been taken into accounts for classification of the next level, but they neglect the word representations residing in category labels. We argue that the word representations of the parent category label into which the text is classified at the upper level will be very helpful for the multi-label classification at the next level due to the fact that the discriminant features for the parent categories may not be discriminable in the subcategories.

In this section, we describe the method used for the joint embedding of parent category labels and textual words to incorporate conceptual inclusion relationship into the text classification process. Word embedding approaches [27, 28] have been widely applied in word representation of texts. In this paper, we use the GloVe embedding tool for word representation of texts. At the preprocessing stage, the label of predicted parent category is extracted for each text and then is combined with the corresponding text. The parent category and text are embedded in the same space.

Let $C^l = \{c_1^l, c_2^l, \ldots, c_{N_l}^l\}$ be the set of category nodes at the *l-th* level, where $c_k^l$ represents the *k-th* category node at the *l-th* level, $N_l$ is the number of category nodes at the *l-th* level, and $C^l \subseteq C$. The category node into which the text is classified and predicted at the *l-th* level can be obtained by mapping the probability distribution from the softmax layer output towards its corresponding category.

Let $T = \{x_i\}_{i=1}^n$ be the set containing *n* texts where each $x_i$ is labeled with their corresponding category labels. Here, Let *L* be the number of levels in the hierarchy $H = (C, \mathrm{E}, \mathrm{r}, \tau)$. We use the notation $C_i^l$ to denote the category node that the text $x_i$ is beforehand labeled with for any text $x_i \in T$. So it is not difficult to find that $C_i^l \in C^l$. The category label of $C_i^l$ is denoted as $\tau(C_i^l)$. In the situation, the set *S* of all possible labels of the text $x_i$ is $S = \{\tau(C_i^k)\}_{k=1}^L = \{\tau(C_i^1), \tau(C_i^2), \ldots, \tau(C_i^L)\}$.

We propose a joint embedding method for the text representation that is obtained by combining the text with its related parent category and further used as the input at each level. For a given text, its input at each level is different because the classification information from the upper level is considered for classification of the next level.

The combination between the text and its related parent category is made by a concatenation operation defined as shown in Equation (1), where $z_{i,j}$ represents the text representation of the text $x_i$ at the $j$-th level, $\tau(C_i^{j-1})$ is the textual label of the category node $C_i^l$, and $\oplus$ is a concatenation operation that concatenates the textual label $\tau(C_i^{j-1})$ and the text $x_i$.

$$z_{i,j} = \begin{cases} x_i, & if\ j = 1 \\ \tau(C_i^{j-1}) \oplus x_i, & if\ 2 \leq j \leq L \end{cases} \quad (1)$$

After that, the new formed text $z_{i,j}$ is further vectorized as the word representation of the input text by using the GloVe based word embeddings for the classification at the $j$-th level.

### 3.4 Hierarchical Fine-Tuning by Parameters Transfer

Hierarchical fine tuning refers to the transition of the training parameters of certain layers from the upper level category towards the corresponding lower level categories layer in the classification model for training [25, 30]. In hierarchical text classification tasks, due to the high correlation between the pre-training and target tasks, hierarchical fine tuning is useful for improving the hierarchical text classification performance by reusing parent training in their subcategory training process.

During training the model, parent category parameters is acquired as initialization parameters for their child subcategory training model, through which prior knowledge can effectively utilize the data in the upper levels to contribute categorization in the lower levels [25, 31]. What strategy we used for hierarchical fine-tuning is that we transfer the parameters of ONLSTM from the upper level to the next level as the initialized parameters of the next-level ONLSTM. This process is repeated from top to down in the hierarchy. Hierarchical fine-tuning by parameters transfer leverages the hierarchical relations between the categories in the hierarchy to tackle the sub-category data sparsity problem [25].

In the following, we discuss which specific training parameters can be transferred to the model training of the next level by the fine-tuning technique. Figure 3 illustrates us the model training and parameters transfer by using fine-tuning technique. During model training in a level-by-level manner, for a given level $j$, we need to find the category label $\tau(C_i^{j-1})$ assigned to text $x_i$. The word embedding matrix of combining the category label and the text is built, indexed and are further concatenated. A pre-training word vector by the GloVe tool for word embeddings, which does not participate in model training for the classification based on our approach. In figure 3, the training parameters $W_j$ of the layers related to dropout, ON-LSTM and global max pooling within the dashed box of at level $j$, will be transferred to the next level as initialization parameters.

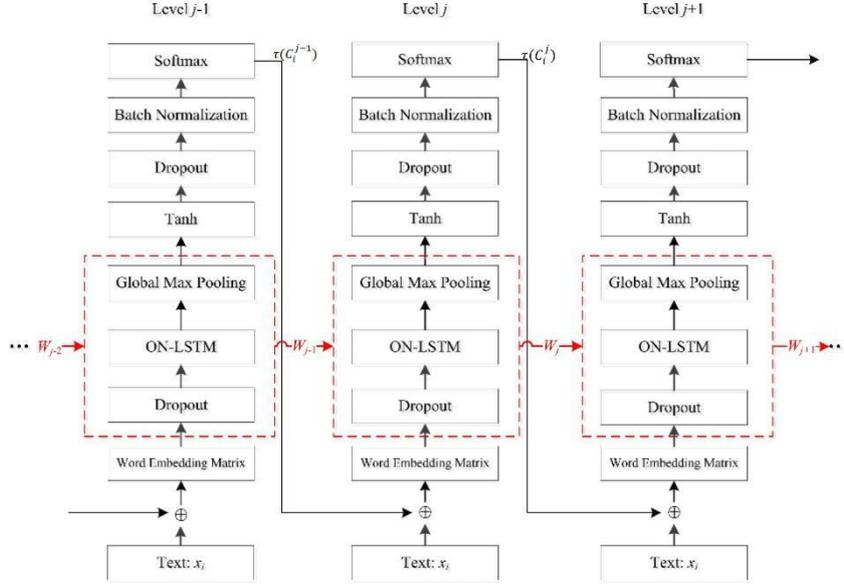

**Fig. 3.** The model training and parameters transfer based on fine-tuning.

### 3.5 The ONLSTM Network Model

Normally, a natural sentence can be presented as a hierarchical structure that we call grammatical information. The ONLSTM model extends the LSTM model and can learn hierarchical structures during the training phase [8]. During training LSTM, updatations between neurons are unrelated and independent from each other. In contrast, ONLSTM adopted a technique of introducing two gates to LSTM units: one is called master input gate represented by $\tilde{\iota}_t$ and the other one is master forget gate represented by $\tilde{f}_t$. For controlling the information which are stored and forgotten based on the state of neurons, a new activation function called **cumax** is used. By implementing such a gate mechanism, the rules of interdependence between neurons are defined, and hence the order and hierarchical differences between neurons can be made.

The structure of ONLSTM neural network model can be also graphically shown in the following Figure 2. After obtaining the text representation $z_{i,j}$ through the above step (equation (1)), we will convert it into semantic vector $w$ through the GloVe based word embedding. In the following, we will use $w_t^j$ to represent all text representations of the *j*-th level label at time *t*. Therefore, we will extract the syntactic structure information in the word vector representation $w_t^j$ through the ONLSTM network model to obtain the text representation $h_t$ for classification based on the softmax function.

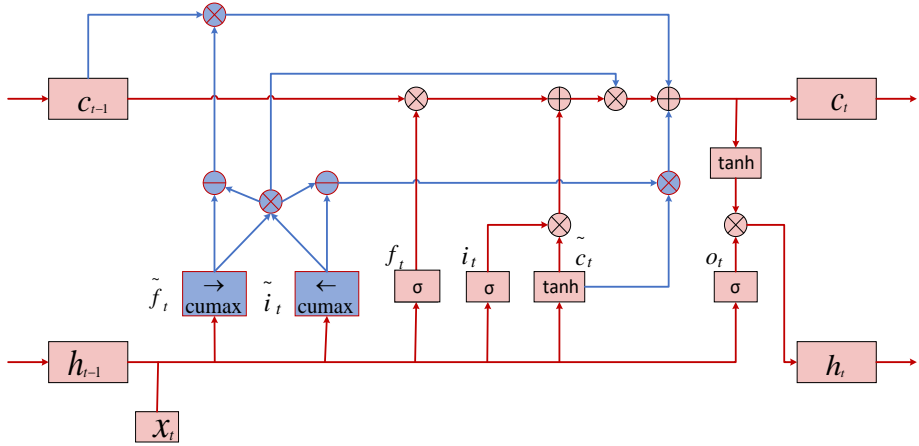

**Fig. 4. The structure of ONLSTM model.**

Based on the ONLSTM neural network, text representation can be obtained by extracting syntactic structure information in the word vector representation $w_t^j$, which can be defined in equation (2).

$$h_t^j = \mathbf{ONLSTM}^j(w_t^j, h_{t-1}^j, W_{on-lstm}^{j-1}) \quad (2)$$

Where $\mathbf{ONLSTM}^j$ is the training process of the ONLSTM layer in the classification model of the *j*-th level label, $h_t^j$ is the hidden state vector of input sequence at time *t*. $W_{onlstm}^{j-1}$ is the weight parameter of ONLSTM network when classifying *j*-1 level categories. The weight parameters of the ONLSTM layer trained on the upper level is transferred to the ONLSTM layer of the adjacent lower level as initialization parameters.

### 3.6 Multi-Layer Perceptron (MLP)

Finally, a two-layer multi-layer perceptron is used to boost the descriptive ability power of neural networks and estimates the probability distribution across classes at level *j*, which is often shown as Equations (3) and (4).

$$d_j = \tanh(W_1 h_t^j + b_1) \tag{3}$$

$$y_j = \mathrm{softmax}(W_2 d_j + b_2) \tag{4}$$

The network parameters are trained for minimizing the cross-entropy of the predicted distributions between true distributions $y$ and $\hat{y}$, which is shown in Equation (5).

$$L(\hat{y}, y) = -\sum_{n=1}^{N} \sum_{c=1}^{C^j} y_n^c \log \hat{y}_n^c \tag{5}$$

Where $C^j$ denotes number of categories at level $j$, $\hat{y}_n^c$ denotes the prediction probabilities, $y_n^c$ is the ground-truth label and the total number of training samples are denoted by $N$.

## 4 Experimental Evaluations and Analysis

### 4.1 Datasets

In order to test the performance of our proposed method, the proposed method is evaluated on two widely used datasets for multi-label classification: Web of Science (WOS) and DBpedia. Web of Science (WOS) dataset [4], consists of 46985 documents, having 7 parent categories with 134 subcategories. Compared to WOS, the DBpedia dataset was first used in [12] for flat text classification. The literature [5] uses the DBpedia ontology to construct a dataset with a three-level taxonomy of classes. The processed DBpedia contains 381,025 documents in total. Comparative details of both datasets are presented in Table 1.

**Table 1.** Description of datasets.

|  | **DBpedia** | **WOS** |
|---|---|---|
| Category Number of Level 1 | 9 | 7 |
| Category Number of Level 2 | 70 | 134 |
| Category Number of Level 3 | 219 | — |
| Total Number of Documents | 381025 | 46985 |

### 4.2 Baselines

In order to validate the classification performance of our method, our method will be compared with the state of the arts of hierarchical classifiers and flat classifiers for multi-label text classification, respectively. These state-of-the-art multi-label classifiers were used as the baselines, which are shown in Table 2.

HFT-ONLSTM: Hierarchical and Fine-Tuning Multi-label Text Classification

The state of the arts of hierarchical multi-label classifiers includes HDLTex [4], HATC [5] and HFT-CNN [25], and these classifiers have reported superior classification performance. On the other hand, the state of the arts of flat classifiers include Bi-directional LSTM with max pooling using a multi-layer perceptron (MLP) [29], Bi-directional LSTM with mean pooling using MLP [33], FastText [34], Structured Self-attentive classifier [35], and BERT [43] using a fully connected MLP.

**Table 2.** Baselines for MHTC.

| Type | Classifier | Reference |
| --- | --- | --- |
| **Flat Classifier** | BiLSTM/MLP/Maxpool | [29] |
| | BiLSTM/MLP/Meanpool | [33] |
| | FastText | [34] |
| | Structured Self-attention | [35] |
| | BERT/MLP | [42] |
| **Hierarchical Classifier** | HDLTex | [4] |
| | HATC | [5] |
| | HFT-CNN | [25] |
| | HFT-ONLSTM (our approach) | |

### 4.3 Hyperparameters Settings

For WOS, a 300 dimensional pre-trained word to vector is trained through GloVe tool as our pre-trained word embeddings. We add 512 hidden units with 0.25 dropout, and fully connected layer having 500 units, with 0.5 dropout, and a batch normalization layer to ONLSTM. For last layer which is a fully connected layer, the number of units in the last layer is set to the category number. The standard Adam optimizer [32] technique is adopted for optimization of all trainable parameters with the initial learning rate of 0.001. If no reduction in the loss value of the test set is observed after 2 epochs, then we reduce the learning rate by 10 times while keeping the batch size at 64. In addition, we adopted the early stopping technique to select the best model. For the DBpedia dataset, all hyperparameters are set the same as WOS, with the exception that the hidden layer size is 300 of ONLSTM.

In this experiment, some hyperparameters such as the dropout parameters, and the numbers of hidden layer units of the ONLSTM network and the fully connected network, are selected based on the comparison results of several rounds of experiments, where we select the best combination of hyperparameters. In addition, we consider the model training problem: the learning rate will often fluctuate in a larger area near the optimal solution if the objective function reaches the convergence state when the fixed learning rate is used. In order to handle this problem, we adopt the learning rate automatic adjustment mechanism, that is, in the early stage of training, the learning rate will be increased to make the training model converge quickly, and then in the late stage (i.e., no validation accuracy improvement is observed) of training, the learning rate will be decreased by 10 times (i.e., one tenth of the initial learning rate) to make the network converge to the optimal solution better.

We implemented the source codes and the datasets based on our approach, and uploaded them to the GitHub website for free downloading and trialing[6].

### 4.4 Evaluation Indexes

The typical evaluation indexes for multi-label text classification are classification performance and complexity.

### 4.4.1 Classification Performance

(1) The classification accuracy is evaluated from two perspectives in this paper. One perspective is to evaluate *the accuracy at each level*, which refers to the classification performance of each level when we provide the true parent category to the classifier while predicting the next category. However, this is not desirable because during inference we should not have access to the correct parent category. So the other perspective is *the overall accuracy* that refers to the accuracy of text classification at the last level, but true parent categories are not provided in advance in the process of classification at each level. Specifically, in the process of hierarchical classification, the required parent categories come from the classification results of the algorithm at the upper level. The flat algorithms only consider classification of the last level, does not deal with the middle levels in the hierarchy, so the result of each flat algorithm is only overall accuracy.

(2) Precision ($P$), Recall ($R$) and F1-score ($F_1$), the performance metrics from information retrieval, are often used for cassification performance analysis [46]. Precision expresses the proportion of the true positive instances to the instance that our model predicts positive. In other words, precision tells us how much we can trust the model when it predicts an example as positive. The recall predictive accuracy of the model for the positive class. Intuitively, it measures the ability of the model to find all the positive instances in the dataset. F1-score aggregates Precision and Recall measures under the concept of harmonic mean.

---

[6]https://github.com/Bonphy/HFT-ONLSTM

(3) The coverage error expresses how many top-scored final prediction labels we have to include without missing any ground truth label [46,47]. This is significant if we want to know the average number of top-scored-prediction required to predict in order to not miss any ground truth label. The *coverage error* (*Cov*) is computed as follows:

$$Cov = \frac{1}{n}\sum_{i=1}^{n} \left( \max_{c_{ij} \in C_i} \{r_i(c_{ij})\} - 1 \right) \tag{6}$$

Where $C_i \subseteq C$ is the categories set of *i-th* instance and $C = \{c_j | j = 1 \cdots N\}$ is the set of all categories. $r_i(c_{ij})$ denotes the rank predicted for $c_{ij}$ and is defined as equation (7).

$$r_i(c_{ij}) = |\{k | \hat{c}_{ik} \geq \hat{c}_{ij}\}| \tag{7}$$

Where $\hat{c}_{ij}$ is the predicted score for $c_{ij}$.

(4) Ranking Loss is defined as the number of times that irrelevant labels are ranked higher than relevant labels [46]. *Ranking Loss* (*RLoss*) can be calculated as shown in Equation (8).

$$RLoss = \frac{1}{n}\sum_{i=1}^{n} \frac{1}{|C_i||\bar{C}_i|} |\{(c_{ij}, c_{ik}) | r_i(c_{ij}) < r_i(c_{ik}), (c_{ij}, c_{ik}) \in C_i \times \bar{C}_i\}| \tag{8}$$

Where $\bar{C}_i$ is the complementary set of $C_i$ with respect to $C$.

### 4.4.2 Model Complexity

Second, classification complexity is calculated by the total number of training parameters that a classifier to be tested is trained over our two benchmark datasets DBpedia and WOS. More training parameters mean that the classifier will consume more time for text classification.

## 4.5 Experimental Results and Analysis

### 4.5.1 Classification Accuracy

*4.5.1.1 Overall Accuracy Comparison with Flat Classifiers*

Here our proposed method is compared with the state-of-the-art flat baseline classifiers including BiLSTM/MLP/Maxpool, BiLSTM/MLP/Meanpool, FastText, Structured Self-attention and BERT/MLP over datasets DBpedia and WOS. The experimental results are shown in Table 3. From the overall accuracy, its means that the classification accuracy of the last level categories of text without providing real parent categories, as classifiers by itself predict parent categories used in the classification process. The overall accuracy based on flat classifiers does not consider the hierarchical structure among categories (all of the categories are considered at the same level).

**Table 3.** The overall accuracy comparison with flat classifiers.

| Flat Classifier [Reference] | Overall Accuracy (%) | |
|---|---|---|
| | DBpedia | WOS |
| BiLSTM/MLP/Maxpool [29] | 93.17 | 77.93 |
| BiLSTM/MLP/Meanpool [33] | 92.67 | 62.92 |
| FastText [34] | 93.36 | 63.66 |
| Structured Self-attention [35] | 93.74 | 71.95 |
| BERT/MLP [42] | **95.17*** | 80.13 |
| HFT-ONLSTM (our approach) | 95.04 | **81.55*** |

From Table 3, we can find that our approach is very competitive against the classification accuracy, and outperforms almost all of the-state-of-the-art flat classifiers over the two datasets DBPedia and WOS, with the exception that the overall accuracy (95.17%) of BERT/MLP is slightly higher than the overall accuracy (95.04%) of our HFT-ONLSTM. The overall accuracy of our HFT-ONLSTM method over WOS reaches 82.62%, which is superior to the other flat classifiers (lower than 78%). Compared to the other classifiers, the performance of flat classifier BiLSTM/MLP/Meanpool and FastText are not very good over both the datasets, particularly in WOS, for multi-label text classification.

What is worth noting is that the flat classifier BERT/MLP has overall illustrated a very promising performance of multi-label text classification although it has a slightly lower overall accuracy (80.13%) in comparison with the overall accuracy (81.55%) of our HFT-ONLSTM. What is surprised to us is that BERT/MLP achieves such a classification performance without considering the extra structural knowledge in the hierarchy. This means that a multi-label text classification approach taking into account hierarchical category structure does not always have superior classification

HFT-ONLSTM: Hierarchical and Fine-Tuning Multi-label Text Classification

performance to the flat text classifiers although the experimental results in Table 3 show that our HFT-ONLSTM has an excellent performance over DBPedia and WOS.

*4.5.1.2 Accuracy in Comparison with Hierarchical Classifiers*

In this section, we compared our proposed method with the multi-label state-of-the-art hierarchical text classifiers, HDLTex, HATC and HFT-CNN. The experiments were still made over the two data sets DBpedia and WOS, respectvely. The experimental results are respectively shown in Tables 4 and 5.

**Table 4.** The accuracy comparison with hierarchical classifiers over DBpedia.

| Hierarchical Classifier [Reference] | Accuracy (%) | | | |
| --- | --- | --- | --- | --- |
| | Level 1 | Level 2 | Level 3 | Overall Accuracy |
| HDLTex [4] | 98.97 | 97.03 | 95.89 | 92.08 |
| HATC [5] | 99.21 | 96.03 | 95.32 | 93.72 |
| HFT-CNN [25] | 98.99 | 96.02 | 93.58 | 93.58 |
| HFT-ONLSTM (our approach) | **99.43*** | **97.46*** | **97.26*** | **95.04*** |

**Table 5.** The accuracy comparison with hierarchical classifiers over WOS.

| Hierarchical Classifier [Reference] | Accuracy (%) | | |
| --- | --- | --- | --- |
| | Level 1 | Level 2 | Overall Accuracy |
| HDLTex [4] | 89.98 | 85.19 | 76.65 |
| HATC [5] | 89.32 | 82.42 | 77.46 |
| HFT-CNN [25] | 89.66 | 79.16 | 79.16 |
| HFT-ONLSTM (our approach) | **90.28*** | **86.14*** | **81.55*** |

In the tables 4 and 5, Level 1, Level 2 and Level 3 respectively correspond to the first, the second and the third level in the hierarchical structure. According to Table 1, the dataset DBpedia has 3 levels, and the dataset WOS has only 2 levels. The accuracy corresponding to each Level $i$ ($1 \leq i \leq 3$) refers to the classification accuracy at the *i-th* level while providing the true category of text to the upper level. The *Overall Accuracy* still refers to the classification accuracy of the last level labels of text without providing true parent categories, similar to the flat classifiers.

We can find that our approach is superior to the hierarchical classifiers HDLTex, HATC and HFT-CNN over the two datasets DBPedia and WOS from Tables 4 and 5.

Specifically speaking, for the classification accuracy at each level of DBPedia and WOS, our HFT-ONLSTM approach significantly outperforms the other three hierarchical methods. Furthermore, the overall accuracy of our method over DBpedia and WOS is also obviously better than the other approaches HDLTex, HATC and HFT-CNN.

The reasonable explanations for the results are as follows. First, the joint embeddings of text and categories are effective and work well in the level-by-level hierarchical classification. For a text to be classified, the label embeddings of its parent category label (at the upper level) are introduced that can improve the classification performance at the next level by concatenating the embeddings of the text and the textual label of the upper parent category. A text belonging to a parental category would be most likely to belong to some subcategory of the parental category. Moreover, the word embedding of parental category labels can effectively help the text learn the classification feature at the lower level. Second, we use the hierarchical fine tuning technique to improve the classification accuracy by taking full advantage of the parent category training information to their subcategories. In this way, the training parameters from corresponding layers in the classification model can be transferred from the upper level category towards the lower level category for training according to the hierarchy of categories. Third, the use of the ONLSTM model also works well in capturing text feature representation, which can establish the renewal rules of interdependence between the neurons and learn hierarchical structures naturally during the training process.

What is worth noting in Tables 4 and 5 is the classifier HFT-CNN. It also uses the fine-tuning technique for hierarchical multi-label text classification, but its performance is almost the lowest among the four hierarchical classifiers. We believe that fine-tuning technique is very helpful in improving the classification accuracy of HFT-CNN by taking full advantage of the training information at the upper level for the training of the next level. An underlying reason why HFT-CNN almost has the lowest performance is that it uses the CNN neural model for learning the features of texts without considering the embedding of the parent categories for the classification at the next level. This also brings about the situation where the accuracy of HFT-CNN at the last level is the same as its overall accuracy. HFT-CNN is originally designed that focuses on HMTC for short texts, which adopts word embedding and the convolutional layer as parameters to learn the current level of the category hierarchy. However, the hierarchical approaches including HDLTex, HATC and HFT-CNN work well in learning a chain structure on the text. The neural network models that they adopt are difficult to deal with the hierarchical relationships between words, and the situation where sentences reside in paragraphs. We believe that the performance of hierarchical text classifiers is most likely to be improved if some neural network models with excellent performance of hierarchical word representations and even hierarchical sentence-level representations.

### 4.5.2 Precision, Recall and F1 Score

In the section, we discuss the experiments and analysis about precision, recall and F1-score of our approach in comparison with some baseline methods described in section 4.2, both for flat and hierarchical classifiers. For coherent experiments and analysis, all the classification approaches have the same reference, i.e., the classification results at the last level. The experimental results of precision, recall and F1-score are shown in Table 6.

**Table 6.** The precision, recall and F1-score comparison.

| Classifier [Reference] | DBpedia | | | WOS | | |
| --- | --- | --- | --- | --- | --- | --- |
| | $P(\%)$ | $R(\%)$ | $F_1(\%)$ | $P(\%)$ | $R(\%)$ | $F_1(\%)$ |
| BiLSTM/MLP/Maxpool [29] | 93.06 | 92.27 | 92.22 | 77.71 | 75.99 | 76.37 |
| BiLSTM/MLP/Meanpool [33] | 92.03 | 91.14 | 91.48 | 62.92 | 60.78 | 61.34 |
| FastText [34] | 92.73 | 91.85 | 92.20 | 64.07 | 61.05 | 61.59 |
| Structured Self-attention [35] | 92.90 | 93.10 | 92.91 | 70.56 | 69.77 | 69.52 |
| BERT/MLP [42] | **94.51*** | 93.92 | 94.11 | 79.67 | 78.53 | 78.47 |
| HDLTex [4] | 92.16 | 92.08 | 91.94 | 76.86 | 76.65 | 76.53 |
| HFT-CNN [25] | 92.98 | 92.47 | 92.68 | 79.34 | 78.04 | 78.29 |
| HFT-ONLSTM (our approach) | 94.25 | **94.19*** | **94.18*** | **81.31*** | **80.79*** | **80.85*** |

It was observed that, our HFT-ONLSTM approach is overall superior to other baseline algorithms. Specifically speaking, for these three metrics, the results about precision, recall and F1-score of HFT-ONLSTM are all optimal to other baseline approaches over the dataset WOS. On the DBpedia, the precision of our model is slightly lower than BERT/MLP, but both recall and F1-score are slightly higher than BERT/MLP. The HFT-ONLSTM: Hierarchical and Fine-Tuning Multi-label Text Classificationother results about precision, recall and F1-score of HFT-ONLSTM are superior to other baseline approaches over the dataset Dbpedia. A possible reason why BERT is higher in precision is the imbalanced sample distribution of the dataset rather than the apporach itself. In conclusion, we can find that our model is optimal in terms of comprehensive classification performance with respect to the above three metrics.

### 4.5.3 Coverage Error and Ranking Loss

In this section, we also introduce multi-label classification evaluation measures based on ranking to further validate the performance of our method. In the process of flat algorithms, we only consider coverage error and ranking loss at the last level because they have no values at middle levels. For the hierarchical models, we not only process the results of the last level, but also collect the labels of each level to build a larger set of labels for comparison. Table 7 records the coverag errors of different models, and Table 8 shows the ranking loss of ones.

**Table 7.** The coverage error comparison.

| Classifier [Reference] | | Coverage Error | |
|---|---|---|---|
| | | DBpedia | WOS |
| **Flat** | BiLSTM/MLP/Maxpool [29] | 1.3060 | 2.7212 |
| | BiLSTM/MLP/Meanpool [33] | 1.3494 | 4.9445 |
| | FastText [34] | 1.2225 | 3.5869 |
| | Structured Self-attention [35] | 1.2358 | 3.2109 |
| | BERT/MLP [42] | 1.2211 | 2.4685 |
| | HFT-CNN (flat) [25] | 1.2241 | 2.5480 |
| | HFT-ONLSTM (flat) | **1.2202*** | **2.3002*** |
| **Hierarchical** | HFT-CNN [25] | 3.5199 | 4.0363 |
| | HFT-ONLSTM (our approach) | **3.4283** | **3.7080** |

For flat algorithms, it is observed that our model is still superior to other approaches. The coverage error and ranking loss of our method are respectively 1.2202 and ranking loss is 0.00101 over the dataset DBpedia. In contrast, the flat method BiLSTM/MLP/Meanpool has highest Coverage error and Ranking Loss over both datasets. The difference of Coverage errors and Ranking losses between the resting flat methods is not obvious. For example, the values of FastText are very close to them. The coverage error is only about 0.002 higher, and the difference of ranking loss is less than 0.00001. On the WOS, the performance of our model is convincingly optimal, where Coverage error and ranking loss are equal to 2.3002 and 0.00978 respectively, which is much better than other models.

For hierarchical model, both our model and HFT-CNN collect labels of all levels for comprehensive comparison, and the results show that the performance of our model outperforms that of the latter.

**Table 8.** The ranking loss comparison.

| Classifier [Reference] | | Ranking Loss | |
|---|---|---|---|
| | | DBpedia | WOS |
| Flat | BiLSTM/MLP/Maxpool [29] | 0.00140 | 0.01294 |
| | BiLSTM/MLP/Meanpool [33] | 0.00160 | 0.02966 |
| | FastText [34] | 0.00102 | 0.01945 |
| | Structured Self-attention [35] | 0.00108 | 0.01662 |
| | BERT/MLP [42] | 0.00102 | 0.01104 |
| | HFT-CNN (flat) [25] | 0.00103 | 0.01164 |
| | HFT-ONLSTM (flat) | **0.00101*** | **0.00978*** |
| Hierarchical | HFT-CNN [25] | 0.00083 | 0.00895 |
| | HFT-ONLSTM (our approach) | **0.00071** | **0.00732** |

According to the experimental results, our model has a higher score to the true labels of each sample, making the labels ranking higher than the incorrect labels. Furthermore, we need to go down the label sort list shorter than the other models so as to cover all the relevant labels of the sample.

### 4.5.4 Analysis for Training Parameters

Generally speaking, a smaller number of training parameters often leads to lower computational complexity and lower computational cost. Hierarchical classifiers often have more training parameters than flat classifiers due to the fact that there are multiple levels in the hierarchy. Hierarchical classifiers need to make classification for each level. The number of training parameters of flat classifiers is close to that of hierarchical classifiers with respect to a single level. The more levels the hierarchy has, the more training parameters hierarchical classifiers need. So in the followings, we will only compare the numbers of number of training parameters among different hierarchical classifiers.

Table 9. Number of training parameters.

| Hierarchical Classifier [Reference] | Number of Training Parameters (million) | | |
|---|---|---|---|
| | DBpedia | WOS | Total Number |
| HDLTex [4] | 3200 | 1800 | 5000 |
| HATC [5] | 21 | 13 | 34 |
| HFT-CNN [25] | 83 | 55 | 138 |
| HFT-ONLSTM (our approach) | 83 | 57 | 140 |

Table 9 shows the comparison of model parameters between the four HMTC models including HDLTex, HATC, HFT-CNN and our HFT-ONLSTM model respectively over the two datasets DBpedia and WOS. We respectively reported the number of training parameters over each dataset and the total number of parameters over both the datasets. The number of training parameters over a given dataset is calculated by summing all the parameters of each level taking part in training over the dataset. The total number of parameters can be obtained by summing all the parameters over DBpedia and WOS.

In Table 9, it is clear that total training parameters of HDLTex are much higher than those of the other three classifiers HATC, HFT-CNN and HFT-ONLSTM, and that HATC has the lowest number of training parameters. The reason is that HDLTex is a local HMTC approach where the number of classifiers is determined through the number of nodes in the hierarchy. HDLTex has to train a neural network model for each hierarchical node, and therefore inevitably produces large volumes of training parameters over the two datasets, approximately reaching to about 5000 million. HATC has a global classifier with an attention mechanism for the whole hierarchy, so it has the lowest training parameters, approximately reaching to 34.

What is interested in is that HFT-CNN and HFT-ONLSTM have a very close number of training parameters, 138 and 140 respectively. As HFT-ONLSTM is a level-by-level HMTC method, the number of classifiers is determined by the number of levels in the hierarchy. However, although local approaches often have more parameters due to more training models required, the number of parameters required in our HFT-ONLSTM approach is close to the number of HATC parameters. It is currently the least number of parameters required in the current local HMTC approaches.

### 4.5.5 Ablation Study on HFT-ONLSTM Approach

In this section, we made an ablation study on the involved components in our HFT-ONLSTM approach. Overall speaking, our HTF-ONLSTM approach is involved in two key components: one is the joint embeddings based on parent category and the text for improving multi-label classification performance, and the other is the fine-tuning technique for handling data sparsity problem. By using joint embeddings of parent category and the text, text classification at each level can consider the classification result at the upper level, which is expected to improve the classification performance. In hierarchical classification, the lower the hierarchical level, the worse the categorization performance. The classification at lower levels is often worse trained due to the lack of data, and thus finely tuned parameters can be transferred to lower levels for handling data sparsity problem. In the work [25], some observations have been obtained that fine-tuning technique indeed can improve hierarchical classification performance no matter how the sample data is represented, so in this paper we will independently consider joint embedding and fine-tuning and not observe the dependency between them. In the ablation study, we observe the experimental results from two perspectives: we just use the ONLSTM model, global max pooling and MLP for hierarchical text classification without using the joint embedding and the fine tuning technique, respectively.

Table 10 is the performance analysis against the joint embedding of parent category and the text. We observed that our HFT-ONLSTM with joint embedding approach can significantly improve the text classification accuracy in comparison with the approach without using joint embedding over both DBpedia and WOS, no matter what the true parent category or the predicted parent category is embedded. What is different is that the accuracy at each level is based on the embedding of true parent categories and the overall accuracy is based on the embedding of predicted categories of the upper level. We also observed that the joint embeddings based on true parent categories have an obviously higher accuracy in comparison with that based on predicted parent categories.

Then, we observe the effect of using fine-tuning technique. We say that the data sparsity problem often exists in hierarchical classification. As the level becomes lower, the number of the training samples per category will become smaller, which makes the model training worse at lower levels due to the lack of training sample. Fine-tuning technique can handle data sparsity by reusing the finely tuned parameters at the upper level in their subcategory training process at the lower level. In this paper, we observe the number of iterative training over the two datasets DBpedia and WOS. The experimental results are shown in Table 11.

**Table 10.** The accuracy analysis against joint embeddings.

|  | DBpedia | | | | WOS | | |
| --- | --- | --- | --- | --- | --- | --- | --- |
|  | Level 1 | Level 2 | Level 3 | Overall | Level 1 | Level 2 | Overall |
| Our Approach without Joint Embedding | 99.43 | 96.79 | 95.13 | 95.13 | 90.28 | 81.93 | 81.93 |
| HFT-ONLSTM (our approach) | 99.43 | 97.46 | 97.26 | 95.04 | 90.28 | 86.14 | 81.55 |

**Table 11.** The performance analysis against fine-tuning.

|  | Number of iterative training | |
| --- | --- | --- |
|  | DBpedia | WOS |
| Our Approach without Fine-Tuning | 10 | 14 |
| HFT-ONLSTM (our approach) | 12 | 9 |

From Table 11, we can observe that the number of iterative training over DBpedia increases by 20 percent when using the fine-tuning technique (from 10 to 12), but the number of iterative training over WOS decreases by about 36 percent. Further analysis can be made over the two datasets. From Table 1, each category in DBpedia contains more than 1730 textual instances on average, but each category in WOS only contains more than 350 textual instances on average. We can obviously find that the model training based on WOS has a very serious data sparsity problem against DBpedia. Especially for the categories at the lowest level, there is lack of enough samples for effective model training. However, what is very interesting is that our HFT-ONLSTM approach takes much less iterations than the approach without using fine-tuning for model convergence over the WOS dataset with data sparsity problem. In contrast to WOS, the DBpedia dataset has no severe data sparsity problem, using the fine-tuning technique even consumes more time of training iterations. So we can empirically validate that our HFT-ONLSTM can more effectively handle data sparsity problem for hierarchical classification than the approach without using fine-tuning technique.

At last, we made a brief summary. From the holistic experiments and analysis mentioned above, we can conclude that the performance of our HFT-ONLSTM method is superior to the state-of-the-art hierarchical and flat multi-label text classification approaches, in particular reducing computational costs while achieving superior performance. The two key components in our approach including joint embedding and fine-tuning not only improve the performance of HMTC, but also can accelerate the convergence of model training with data sparsity problem.

## 5  Conclusion

In this paper, we proposed a local hierarchical multi-label text classifier which employs both the joint embedding of text and parent categories and the hierarchical fine-tuning technique for effective and efficient HTMC. Extensive experiments were made over two benchmark datasets in comparison with the state-of-the-art approaches including flat and hierarchical multi-label text classifications, and the results show that our HFT-ONLSTM approach outperforms these approaches, in particular reducing computational costs while achieving superior performance. We can also find some meaningful insights into these certain convincing results. First, it seems that both word representation of category labels and hierarchical semantics are very helpful in improving the performance of HMTC. Second, learning the joint embeddings based on parent categories and textual data can make the text more discriminable in the subcategories and therefore can relieve the error propagates from the higher levels to the lower ones. Third, our approach adopting the fine-tuning technique can accelerate the convergence of model training with data sparsity problem.

In the future work, we will study the HMTC classification method with more complex situations, e.g., directed acyclic graph based HTMC, and model compression based deep neural network optimization for HTMC.

**Acknowledgments**

This work was partially supported by the National Key R&D Program of China under Grants (2018YFC0830605).

**Conflict of Interest**

The authors declare that they have no known competing financial interests or personal relationships that could have appeared to influence the work reported in this article.

They have no conflicts of interest to declare that are relevant to the content of this article.